\documentclass[a4paper]{article}
\usepackage{spconf}
\usepackage{graphicx}
\usepackage[hyphens]{url}
\usepackage{algorithm}
\title{Radiology Guideline-derived Decision Graph for \\Efficient Clinical Decision Support}
\name{Xuewen Zhou, Zhen Chen$^{\dag}$\thanks{ $^\dag$Corresponding author.}}
\address{Department of Data Science and Artificial Intelligence, The Hong Kong Polytechnic University}

\begin{document}
\maketitle
\raggedbottom
\begin{abstract}
Radiological risk stratification and management planning are essential downstream tasks in medical imaging. Automating these tasks is challenging because clinical guideline criteria are distributed across prose, tables, and cross-referenced definitions. We present Guideline-derived Decision Graph (GDGraph), a framework that decouples guideline compilation from case evaluation. During compilation, a large language model partitions guidelines into section-based chunks, extracts local decision subgraphs with linked definitions, and aggregates related branches into an executable decision graph. During inference, a fast System One model answers localized clinical questions while deterministic traversal resolves Boolean logic, turning opaque single-pass predictions into a traceable decision path without task-specific training. Across 347 cases spanning renal cyst, adnexal mass, and thyroid nodule guidelines, GDGraph improves System One accuracy from 57.3\% to 90.8\%, outperforming retrieval-augmented GPT-4o (80.1\%) and graph-guided GPT-4o with chain-of-thought (82.7\%) at substantially lower latency and cost. The source code will be available.
\end{abstract}

\begin{keywords}
Clinical guidelines, decision graphs, radiological decision support, System One models
\end{keywords}

\section{Introduction}
Risk stratification and management planning are essential downstream tasks in medical imaging that connect radiological findings to clinical care \cite{an2019birads}. Clinical guidelines standardize these decisions across tasks such as renal cyst management \cite{richard2023cua}, adnexal mass assessment \cite{strachowski2023orads}, and thyroid nodule stratification \cite{ha2021ktirads}, where final recommendations follow from explicit, discrete decision criteria. Automating these tasks therefore requires systematically following the interdependent decision steps specified in each guideline, progressing from case-specific findings through successive criteria to a final category or recommendation.

Pretrained language models can process free-text clinical reports and imaging findings directly through prompting without task-specific training \cite{thirunavukarasu2023large, HU2024105321}, yet have been shown to frequently deviate from established diagnostic guidelines \cite{hager2024evaluation}. Chain-of-thought (CoT) prompting helps these models with multi-step clinical problems by eliciting reasoning steps in text before a conclusion \cite{wei2022chain}. In guideline-based risk stratification and management, clinical criteria and conditional rules also provide a structured basis for decomposing the overall task into a series of localized decisions, such as verifying a specific imaging feature or comparing a measurement against a threshold. System One models such as Jev \cite{almeida2026systemone} offer a suitable interface for these localized decisions by directly scoring discrete candidate options without generating intermediate text . Thus, the primary challenge is to organize these local judgments into an explicit decision process that preserves the guideline's conditional dependencies and yields a traceable outcome for high-stakes decisions~\cite{rudin2019stop}.

Existing clinical frameworks incorporate guideline knowledge in different ways. For a given patient case, retrieval-augmented generation (RAG) systems \cite{lewis2020rag} such as accGPT \cite{rau2023context} retrieve relevant medical passages for the language model to interpret during response generation. CPGPrompt \cite{deng2026cpgprompt} converts segmented guideline text into decision trees for stepwise evaluation, with its reported workflow relying on manual selection of relevant text sections and manual validation of the generated trees. Such decision structures could harness the capabilities and efficiency of System One language models to support practical applications in clinical decision-making. However, automatic construction of these structures from raw guideline documents remains challenging because it requires integrating tabulated criteria and thresholds with terminology definitions and exception rules distributed across the document.

\begin{figure*}[t]
\centering
\includegraphics[width=\textwidth]{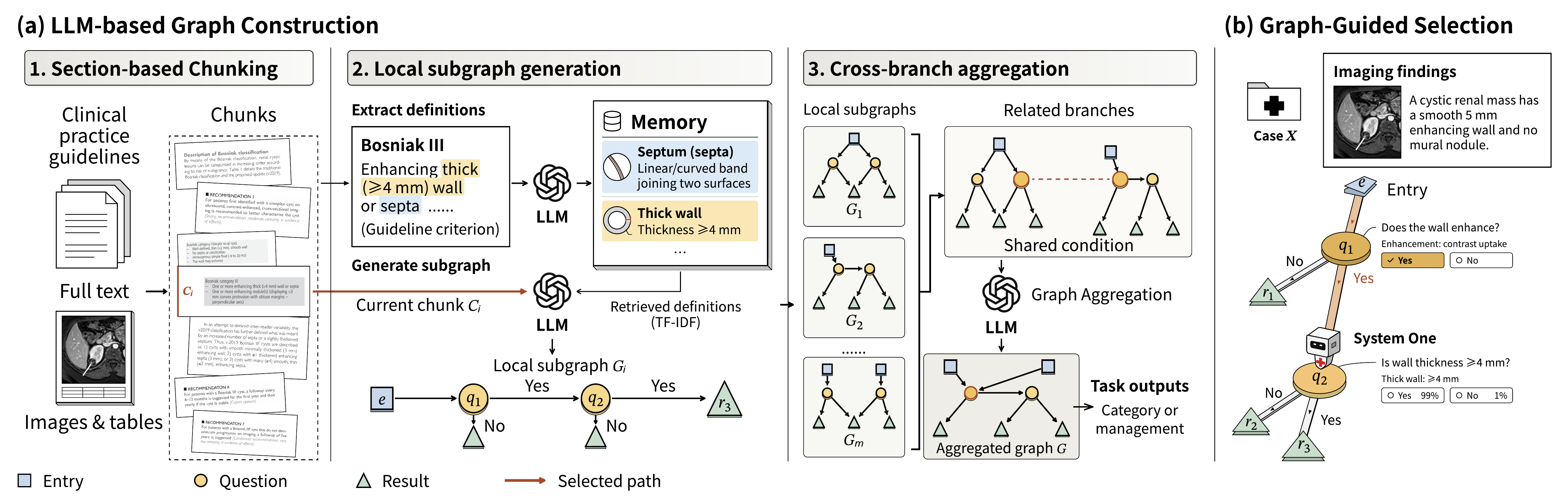}
\caption{Overview of the GDGraph framework. \textbf{(a) LLM-based Graph Construction:} Offline, an LLM partitions clinical guidelines into $m$ section-based chunks $\{C_i\}_{i=1}^{m}$, extracts local subgraphs $\{G_i\}_{i=1}^{m}$ with TF-IDF-retrieved definitions from a global memory, and aggregates related branches into a unified decision graph $G$. \textbf{(b) Graph-Guided Selection:} Online, a System One model answers localized questions using case findings and attached definitions, guiding traversal along the selected path to reach a traceable decision.}
\label{fig:overview}
\end{figure*}

To address this, we present Guideline-derived Decision Graph (GDGraph), an automated framework that compiles raw clinical guidelines into executable decision graphs for fast, traceable System One inference (Fig.~\ref{fig:overview}). By decoupling offline LLM-based guideline structuring from online deterministic traversal, GDGraph reduces multi-step classification to localized node-level judgments while recording an explicit decision trace. Across 347 radiology cases spanning three guidelines, GDGraph raises System One accuracy from 57.3\% to 90.8\%, outperforming unstructured RAG (80.1\%) and graph-guided GPT-4o with CoT (82.7\%) at substantially lower latency and cost. These results suggest the practical value of GDGraph for clinical decision support.

\section{Methodology}
\label{sec:methods}
\subsection{Overview}
\label{sec:overview}
Fig.~\ref{fig:overview} illustrates the proposed Guideline-derived Decision Graph (GDGraph) framework, which decouples offline guideline compilation from online case evaluation. In the offline stage (Fig.~\ref{fig:overview}(a); Algorithm~\ref{alg:gdgraph}), given a clinical practice guideline $D$ containing narrative text, figures, and tables, an LLM-based pipeline partitions $D$ into $m$ section-based chunks $\{C_i\}_{i=1}^{m}$, extracts local subgraphs $\{G_i\}_{i=1}^{m}$ grounded in a global definition memory $M$ via Term Frequency–Inverse Document Frequency (TF-IDF) retrieval \cite{sparck1972statistical,salton1988term}, and aggregates related branches across sections into a unified decision graph $G = (\mathcal{S}, \mathcal{A}, \mathcal{E}, \mathcal{Y})$. In the online stage (Fig.~\ref{fig:overview}(b)), given the imaging findings of a clinical case $X$, a fast System One model answers localized questions at active nodes using attached definitions, while deterministic traversal resolves Boolean transitions along the selected path to reach a traceable category or management decision.

\subsection{Guideline-derived Decision Graph}
\label{sec:construction}

\noindent\textbf{Graph Representation.} The decision graph $G = (\mathcal{S}, \mathcal{A}, \mathcal{E}, \mathcal{Y})$ comprises clinical states $\mathcal{S}$ (clinical contexts or intermediate conclusions, with entry states $\mathcal{S}_0 \subset \mathcal{S}$), atomic conditions $\mathcal{A}$, directed transition rules $\mathcal{E}$, and candidate outcomes $\mathcal{Y}$ (diagnostic categories or management actions). Each state and outcome carries a clinical label, a source-derived definition, and supporting text spans from $D$. Each atomic condition $a \in \mathcal{A}$ represents a single traceable criterion, written as \mbox{$a = (\tau_a, \kappa_a, \mathcal{D}_a)$}, where $\tau_a$ specifies the clinical feature and its anatomical or temporal scope (e.g., lesion diameter), $\kappa_a$ denotes the comparison operator and value (e.g., $< 10\,\mathrm{cm}$ or presence/absence), and $\mathcal{D}_a$ links to the corresponding definitions in $M$. Deterministic templates render each $a$ and its referenced definitions $\mathcal{D}_a$ into a self-contained question for model evaluation.

Directed rules $e \in \mathcal{E}$ take the form \mbox{$e = (s_e, \phi_e, T_e, \pi_e)$} and connect source states to targets through Boolean combinations of atomic conditions. Here, $s_e \in \mathcal{S}$ is the prerequisite state, $\phi_e$ is a logical expression over $\mathcal{A}$ using standard operators ($\land, \lor, \neg$, and $\mathrm{AtLeast}_k$), $T_e \subseteq \mathcal{S} \cup \mathcal{Y}$ denotes the target states or outcomes, and $\pi_e$ indicates whether the rule directly determines, supports, or rules out $T_e$. For instance, requiring both imaging pattern $P$ and diameter below $10\,\mathrm{cm}$ for the same lesion is formulated as $\phi_e = a_P \land a_{<10}$.

\begin{algorithm}[tbp]
\caption{GDGraph Construction}
\label{alg:gdgraph}
\small
\textbf{Input:} Clinical guideline document $D$\par
\textbf{Output:} Decision graph $G = (\mathcal{S}, \mathcal{A}, \mathcal{E}, \mathcal{Y})$\par
\smallskip
\begin{tabular}{@{}r@{\quad}p{\dimexpr\columnwidth-3em\relax}@{}}
1: & \raggedright Partition $D$ into section-based chunks $C = \{C_1, \dots, C_m\}$ \tabularnewline
2: & \raggedright Build definition memory $M \leftarrow \bigcup_{i=1}^m \mathrm{ExtractTerms}(C_i)$ \tabularnewline
3: & \raggedright \textbf{for} each chunk $C_i \in C$ \textbf{do} \tabularnewline
4: & \raggedright \quad Retrieve relevant definitions $d_i \subset M$ via TF-IDF \tabularnewline
5: & \raggedright \quad Extract subgraph $G_i = (\mathcal{S}_i, \mathcal{A}_i, \mathcal{E}_i, \mathcal{Y}_i)$ from $(C_i, d_i)$ \tabularnewline
6: & \raggedright \quad Render questions for each $a \in \mathcal{A}_i$ with $\mathcal{D}_a \subset M$ \tabularnewline
7: & \raggedright \textbf{end for} \tabularnewline
8: & \raggedright Group $\{G_i\}_{i=1}^m$ into overlapping clusters $\mathcal{B} = \{B_1, \dots, B_L\}$ \tabularnewline
9: & \raggedright \textbf{for} each cluster $B_l \in \mathcal{B}$ sequentially \textbf{do} \tabularnewline
10: & \raggedright \quad Load latest updated subgraphs in $B_l$ \tabularnewline
11: & \raggedright \quad Retrieve relevant definitions $d_l \subset M$ via TF-IDF \tabularnewline
12: & \raggedright \quad Merge equivalent states and atoms across $B_l$ \tabularnewline
13: & \raggedright \quad Update affected rules $e \in \bigcup_{G_i \in B_l} \mathcal{E}_i$ ($\phi_e$ and $T_e$) \tabularnewline
14: & \raggedright \textbf{end for} \tabularnewline
15: & \raggedright Assemble unified graph $G = (\mathcal{S}, \mathcal{A}, \mathcal{E}, \mathcal{Y})$ \tabularnewline
16: & \raggedright \textbf{return} $G$ \tabularnewline
\end{tabular}
\end{algorithm}

\noindent\textbf{Section-Based Chunking and Local Subgraph Generation.} We split $D$ by document sections into chunks $C = \{C_1, \dots, C_m\}$ while keeping tables and footnotes intact. An LLM scans all chunks to build a global definition memory $M$ of clinical terms, aliases, and definitions. For each chunk $C_i$, we retrieve relevant definitions $d_i \subset M$ via TF-IDF similarity while keeping explicitly cited definitions, and prompt the LLM to extract states $\mathcal{S}_i$, atomic conditions $\mathcal{A}_i$, and rules $\mathcal{E}_i$ into a local subgraph $G_i$. Named imaging patterns are kept as single conditions linked to $M$, additional criteria such as numerical thresholds are separated into individual conditions, and negative findings are denoted by logical negation ($\neg$).

\noindent\textbf{Cross-Branch Aggregation.} To connect related rules across sections, local subgraphs $\{G_i\}_{i=1}^{m}$ are merged into a unified graph $G$. Guided by TF-IDF similarity, an LLM groups subgraphs sharing clinical features or exceptions into potentially overlapping sets $\mathcal{B} = \{B_1, \dots, B_L\}$, processed sequentially on the latest graph. Within each group $B_l$, the LLM uses the source text and definitions $d_l$ to merge nodes that share identical clinical meaning and judgment criteria, while keeping non-equivalent thresholds separate. When the guideline specifies upgrading or exception clauses, the LLM is instructed to update the conditions ($\phi_e$) and targets ($T_e$) of affected rules $e$, while keeping other valid standalone paths unchanged.

\subsection{Graph-Guided Selection}
\label{sec:selection}

Prior to evaluation, an LLM maps source-supported conclusions in $\mathcal{S} \cup \mathcal{Y}$, including intermediate states and terminal outcomes, to the target task's standardized answer options, as illustrated in Fig.~\ref{fig:overview}(b). During online inference, traversal starts at $\mathcal{S}_0$. At each active state, the System One model answers the questions of outgoing atomic conditions based on the reported imaging findings and attached definitions $\mathcal{D}_a$, while program logic evaluates $\phi_e$ to advance along satisfied branches. The final prediction is derived from the task-relevant conclusions reached during traversal, with the clinical case interpreted against static option definitions. The traversal log records each evaluated question and its answer, preserving the traversed edges as a traceable decision path.

\section{Experiments}
\label{sec:results}

\subsection{Experimental Setup}
\label{sec:setup}

\noindent\textbf{Dataset and Tasks.} We evaluate GDGraph on 347 text-based radiology cases curated from PMC-Patients-V2 \cite{zhao2023pmc} and PubMed Central case reports across three clinical guidelines (Table~\ref{tab:datasets}). Cases were screened for guideline-relevant imaging findings and clinical context. Reference labels were taken from explicit source labels or derived by applying guideline criteria to the reported findings. Case descriptions were reformulated using GPT-6.1-sol to remove explicit diagnostic labels and reduce the risk of answer leakage. CUA evaluates management selection given the clinical context, whereas O-RADS and K-TIRADS assign ultrasound risk categories from imaging descriptions.

\begin{table}[tbp]
\centering
\caption{Summary of evaluated clinical guidelines, target tasks, and case counts.}
\label{tab:datasets}
\small
\setlength{\tabcolsep}{4pt}
\renewcommand{\arraystretch}{1.1}
\begin{tabular*}{\columnwidth}{@{\extracolsep{\fill}}llr@{}}
\hline
\textbf{Guideline} & \textbf{Clinical Task} & \textbf{Cases} \\
\hline
CUA (2023) \cite{richard2023cua} & Renal cyst management & 127 \\
O-RADS US v2022 \cite{strachowski2023orads} & Adnexal mass stratification & 117 \\
K-TIRADS (2021) \cite{ha2021ktirads} & Thyroid nodule stratification & 103 \\
\hline
\textbf{Total} & & \textbf{347} \\
\hline
\end{tabular*}
\end{table}

\noindent\textbf{Implementation and Baselines.} Decision graphs are constructed offline using GPT-6-Luna with low reasoning effort, processing extracted text alongside PDF page images to preserve table structures and footnotes. For online evaluation, we use the System One model Jev (v1.13.0) \cite{almeida2026systemone} and GPT-4o across six settings: (1) baselines without retrieval or GDGraph (\emph{Direct Jev}, \emph{Direct GPT-4o}, and \emph{GPT-4o CoT}), which receive only the clinical case, candidate options, and option definitions; (2) retrieval-augmented generation (\emph{RAG+GPT-4o CoT}), where GPT-4o generates three search queries per case and MedCPT \cite{jin2023medcpt} retrieves the top-$K$ ($K=5$) guideline chunks per query for CoT reasoning; and (3) graph-guided inference (\emph{GDGraph+Jev} and \emph{GDGraph+GPT-4o CoT}), which traverse the same constructed GDGraph for each task.

\noindent\textbf{Evaluation Metrics.} For graph quality (Table~\ref{tab:graph-quality}), we manually verify \emph{node precision} (proportion of atomic conditions with source-supported questions and definitions) and \emph{rule precision} (proportion of extracted transition rules supported by the source guideline). For decision performance and efficiency, we report case accuracy (Acc.), weighted F1 (WF1), and Matthews correlation coefficient (MCC) \cite{opitz2024closer}, alongside mean online API time and cost per case.

\begin{table}[bp]
\centering
\caption{Precision of constructed decision graphs against source guidelines, reported as supported/total (\%).}
\label{tab:graph-quality}
\small
\setlength{\tabcolsep}{4pt}
\begin{tabular*}{\columnwidth}{@{\extracolsep{\fill}}lcc@{}}
\hline
\textbf{Guideline} & \textbf{Node Prec.} & \textbf{Rule Prec.} \\
\hline
CUA & 137/137 (100.0\%) & 51/55 (92.7\%) \\
O-RADS & 136/137 (99.3\%) & 92/98 (93.9\%) \\
K-TIRADS & 216/216 (100.0\%) & 123/123 (100.0\%) \\
\hline
\textbf{Overall} & \textbf{489/490 (99.8\%)} & \textbf{266/276 (96.4\%)} \\
\hline
\end{tabular*}
\end{table}

\label{sec:results}

\begin{table*}[!t]
\centering
\begin{minipage}[t]{\columnwidth}
\vspace{0pt}
\centering
\caption{Comparison of decision performance across 347 radiology cases. Acc.: case accuracy; WF1: weighted F1-score; MCC: Matthews correlation coefficient. Best results within each task are shown in bold.}
\label{tab:main-results}
\small
\setlength{\tabcolsep}{3pt}
\renewcommand{\arraystretch}{1.0}
\begin{tabular*}{\linewidth}{@{\extracolsep{\fill}}lrrr@{}}
\hline
\textbf{Method} & Acc. $\uparrow$ (\%) & WF1 $\uparrow$ (\%) & MCC $\uparrow$ \\
\hline
\multicolumn{4}{@{}l}{\textbf{CUA} ($n=127$)} \\
Direct Jev & 61.4 & 64.1 & 0.555 \\
\textbf{GDGraph+Jev} & \textbf{93.7} & \textbf{94.4} & \textbf{0.922} \\
\cline{1-4}
Direct GPT-4o & 46.5 & 54.4 & 0.412 \\
GPT-4o CoT & 52.8 & 59.3 & 0.447 \\
RAG+GPT-4o CoT & 86.6 & 88.0 & 0.836 \\
GDGraph+GPT-4o CoT & 89.0 & 87.5 & 0.862 \\
\hline
\multicolumn{4}{@{}l}{\textbf{O-RADS} ($n=117$)} \\
Direct Jev & 72.6 & 70.2 & 0.587 \\
\textbf{GDGraph+Jev} & \textbf{92.3} & \textbf{93.1} & \textbf{0.884} \\
\cline{1-4}
Direct GPT-4o & 69.2 & 70.4 & 0.552 \\
GPT-4o CoT & 57.3 & 59.7 & 0.458 \\
RAG+GPT-4o CoT & 84.6 & 85.4 & 0.786 \\
GDGraph+GPT-4o CoT & 78.6 & 82.9 & 0.713 \\
\hline
\multicolumn{4}{@{}l}{\textbf{K-TIRADS} ($n=103$)} \\
Direct Jev & 35.0 & 34.6 & 0.235 \\
\textbf{GDGraph+Jev} & \textbf{85.4} & \textbf{85.6} & \textbf{0.794} \\
\cline{1-4}
Direct GPT-4o & 47.6 & 49.3 & 0.342 \\
GPT-4o CoT & 44.7 & 47.8 & 0.238 \\
RAG+GPT-4o CoT & 67.0 & 68.4 & 0.515 \\
GDGraph+GPT-4o CoT & 79.6 & 80.1 & 0.714 \\
\hline
\end{tabular*}
\end{minipage}\hfill
\begin{minipage}[t]{\columnwidth}
\vspace{0pt}
\centering
\makeatletter
\def\@captype{figure}
\makeatother
\includegraphics[width=\linewidth]{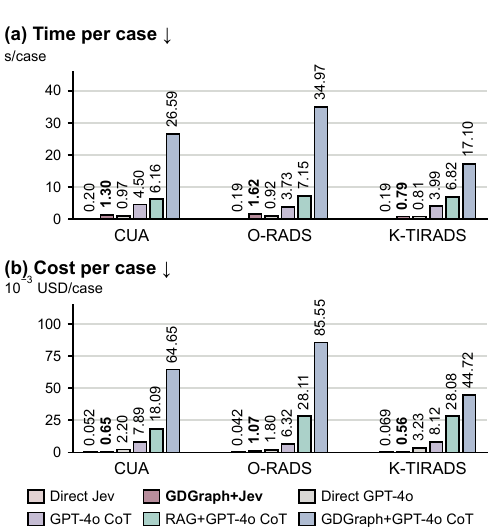}
\caption{Comparison of online inference efficiency across the three guideline tasks, measured by \textbf{(a)} mean cumulative API time (s/case) and \textbf{(b)} mean API cost ($10^{-3}$~USD/case).}
\label{fig:efficiency}
\label{fig:efficiency}
\end{minipage}
\end{table*}

\subsection{Graph Fidelity and Comparison with Retrieval}
In the source-based review (Table~\ref{tab:graph-quality}), the constructed graphs achieve 99.8\% overall node precision and 96.4\% rule precision, showing high source fidelity. On downstream tasks (Table~\ref{tab:main-results}), unguided \emph{Direct GPT-4o} and \emph{GPT-4o CoT} fall below 60\% accuracy on CUA and K-TIRADS. Unstructured retrieval (\emph{RAG+GPT-4o CoT}) raises overall GPT-4o CoT accuracy from 51.9\% to 80.1\%, while graph guidance (\emph{GDGraph+GPT-4o CoT}) reaches 82.7\%. While RAG performs better on O-RADS (84.6\% vs.\ 78.6\%), graph guidance excels on K-TIRADS (79.6\% vs.\ 67.0\%), where stratification jointly depends on nodule composition, echogenicity, and suspicious features, showing that explicit paths with linked definitions help resolve multi-feature criteria.

\subsection{System One Performance and Inference Efficiency}
We next evaluate whether pairing GDGraph with the System One model Jev improves decision accuracy (Table~\ref{tab:main-results}) and compare its online inference efficiency against the GPT-4o baselines (Fig.~\ref{fig:efficiency}). Given only the case content, the candidate options, and their definitions (\emph{Direct Jev}), Jev reaches 57.3\% overall accuracy (35.0\% on K-TIRADS). When the same cases are evaluated using GDGraph (\emph{GDGraph+Jev}), performance improves across all three tasks, reaching 90.8\% overall accuracy and achieving the highest Acc., WF1, and MCC in each guideline. Additionally, \emph{GDGraph+Jev} achieves higher accuracy than both \emph{GDGraph+GPT-4o CoT} (82.7\%) and \emph{RAG+GPT-4o CoT} (80.1\%), while requiring 0.79--1.62\,s per case in online API time (vs.\ 6.16--7.15\,s for \emph{RAG+GPT-4o CoT} and 17.10--34.97\,s for \emph{GDGraph+GPT-4o CoT}) and $0.56$--$1.07 \times 10^{-3}$\,USD per case in API cost (Fig.~\ref{fig:efficiency}). These results indicate that organizing guidelines into localized node judgments and explicit rules enables the System One model to resolve multi-condition criteria with low online inference latency and cost.

\subsection{Error Analysis}
Among the remaining errors, we identify two recurring patterns that highlight the main limitations of GDGraph-guided inference. In 11 K-TIRADS cases, ``ill-defined margins'' or ``irregular shape'' were interpreted as ``irregular margins,'' activating a higher-risk branch because node definitions did not explicitly contrast these terms. In three O-RADS cases, errors arose when mapping qualitative vascularity descriptions (e.g., moderate flow) to node-level criteria. These observations highlight the importance of normalizing radiological descriptions and aligning imaging concepts with model representations across varied reports.

\section{Conclusion}
\label{sec:conclusion}
We proposed GDGraph, a framework that converts radiology guidelines into executable decision graphs for risk stratification and management planning. During compilation, an LLM constructs the graph through section-based chunking, local subgraph extraction, and cross-branch aggregation. During inference, the System One model answers localized questions along the graph, yielding a traceable decision path. Experiments on three guidelines show that GDGraph+Jev achieves higher accuracy than CoT-based GPT-4o baselines at lower latency and cost, indicating its practical value for auditable clinical decision support.

\section{Compliance with Ethical Standards}
This research study was conducted retrospectively using human subject data made available in open access by the PubMed Central Open Access Subset. Ethical approval was not required, as confirmed by the license attached to the open access data.

\section{Acknowledgments}
This work is supported by PolyU Start-up Fund (P0060371). 

\bibliographystyle{IEEEbib}
\bibliography{references}

\end{document}